\documentclass[sigconf, screen]{acmart}
\renewcommand\footnotetextcopyrightpermission[1]{} 
\AtBeginDocument{%
  }

\acmSubmissionID{4855}

\usepackage{cuted}
\usepackage{caption}
\usepackage{dblfloatfix} 
\begin{document}

\title{OVS-DINO: Open-Vocabulary Segmentation via Structure-Aligned SAM-DINO with Language Guidance}

\author{Haoxi Zeng}
\authornote{Both authors contributed equally to this research.}
\email{zenghaoxi@tongji.edu.cn}
\orcid{0009-0005-9575-8766}
\author{Qiankun Liu}
\authornotemark[1]
\email{tjdxlqk@tongji.edu.cn}
\orcid{0009-0003-8471-7569}
\affiliation{%
  \institution{Tongji University}
  \city{Shanghai}
  \country{China}
}

\author{Yi Bin}
\authornote{Corresponding author.}
\affiliation{%
  \institution{Tongji University}
  \city{Shanghai}
  \country{China}}
\email{yi.bin@hotmail.com}

\author{Haiyue Zhang}
\affiliation{%
  \institution{Tongji University}
  \city{Shanghai}
  \country{China}
}

\author{Yujuan Ding}
\affiliation{%
 \institution{Hong Kong Polytechnic University}
 \city{Hong Kong}
 \country{China}}

\author{Guoqing Wang}
\affiliation{%
  \institution{University of Electronic Science and Technology of China}
  \city{Chengdu}
  \country{China}}

\author{Deqiang Ouyang}
\affiliation{%
  \institution{Chongqing University}
  \city{Chongqing}
  \country{China}}

\author{Heng Tao Shen}
\affiliation{%
  \institution{Tongji University}
  \city{Shanghai}
  \country{China}}

\renewcommand{\shortauthors}{}

\begin{abstract}
Open-Vocabulary Segmentation (OVS) aims to segment image regions beyond predefined category sets by leveraging semantic descriptions. While CLIP based approaches excel in semantic generalization, they frequently lack the fine-grained spatial awareness required for dense prediction. Recent efforts have incorporated Vision Foundation Models (VFMs) like DINO to alleviate these limitations. However, these methods still struggle with the precise edge perception necessary for high fidelity segmentation. In this paper, we analyze internal representations of DINO and discover that its inherent boundary awareness is not absent but rather undergoes progressive attenuation as features transition into deeper transformer blocks. To address this, we propose OVS-DINO, a novel framework that revitalizes latent edge-sensitivity of DINO through structural alignment with the Segment Anything Model (SAM). Specifically, we introduce a Structure-Aware Encoder (SAE) and a Structure-Modulated Decoder (SMD) to effectively activate boundary features of DINO using SAM’s structural priors, complemented by a supervision strategy utilizing SAM generated pseudo-masks.  Extensive experiments demonstrate that our method achieves \textbf{state-of-the-art} performance across multiple weakly-supervised OVS benchmarks, improving the average score by 2.1\% (from 44.8\% to 46.9\%). Notably, our approach significantly enhances segmentation accuracy in complex, cluttered scenarios, with a gain of 6.3\% on Cityscapes (from 36.6\% to 42.9\%). 
\end{abstract}

\ccsdesc[500]{Computing methodologies~Image segmentation}
\ccsdesc[300]{Computing methodologies~Information extraction}

\keywords{Open-Vocabulary Segmentation, DINO, Cross-Modal Alignment}


\maketitle
\pagestyle{plain}
\section{Introduction}

\begin{figure}[t]
  \centering
  \includegraphics[width=0.5\textwidth]{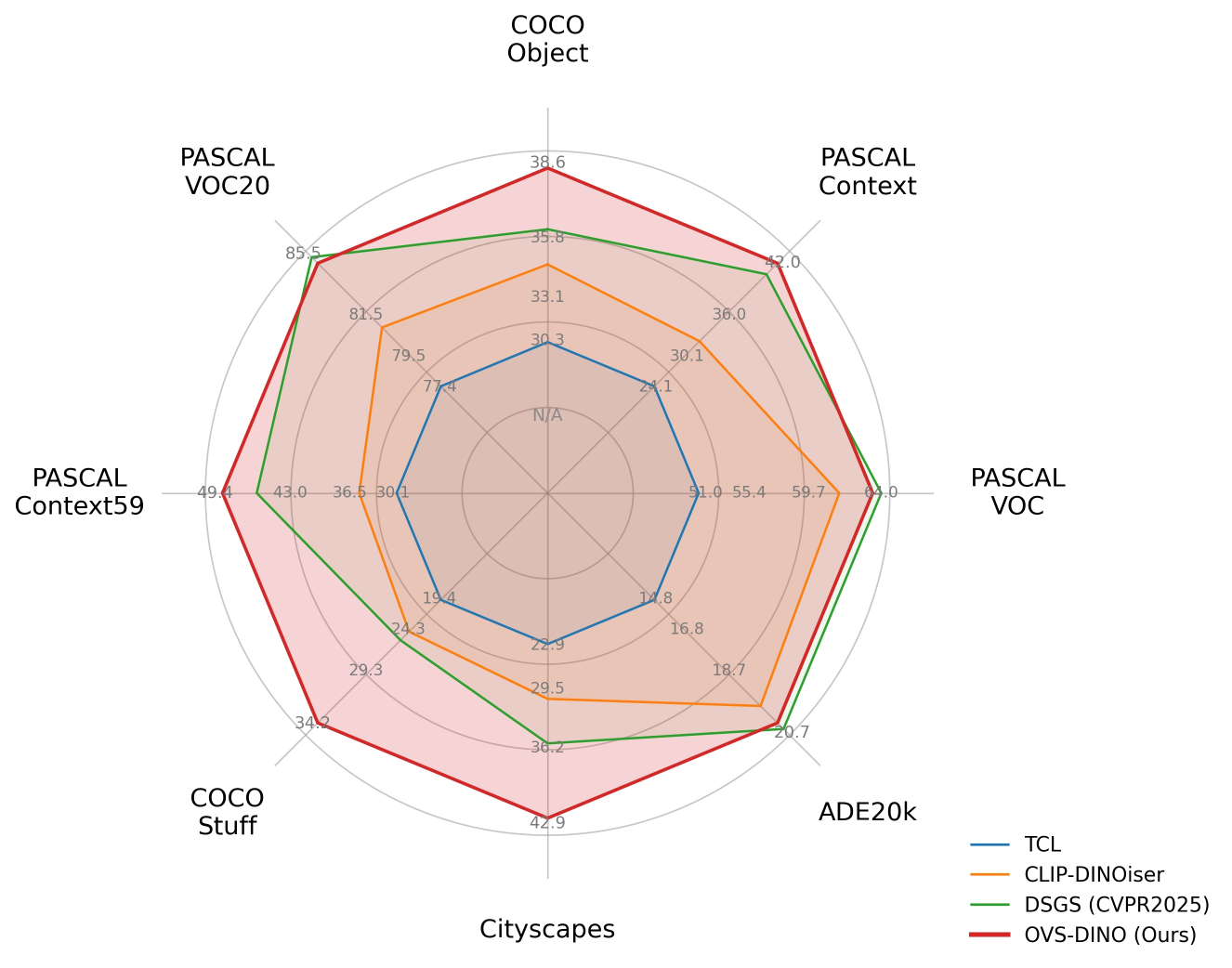}
  \caption{\textbf{Overall performance comparison.} We evaluate our proposed method against state-of-the-art weakly-supervised approaches. The results illustrate that while existing methods struggle with fine-grained segmentation on challenging datasets such as Cityscapes ~\cite{cordts2016cityscapes} and COCO-Stuff ~\cite{caesar2018coco}, our OVS-DINO consistently achieves superior performance across all benchmarks, with particularly significant improvements on complex scenarios.}
  \label{fig:performance}
\end{figure}

Open-Vocabulary Segmentation (OVS) ~\cite{ding2022open,dou2025geopurify,xu2023open,ren2023viewco,mukhoti2023open} aims to overcome the limitations of predefined category sets, enabling semantic-driven image region segmentation over unrestricted vocabularies. Benefiting from strong cross-modal alignment and zero-shot generalization, Vision-Language Models (VLMs) have become a key driver for advancing OVS. The development of VLM pre-training methods, such as ALIGN ~\cite{jia2021scaling} and CLIP ~\cite{radford2021learning}, has demonstrated a remarkable ability to align textual descriptions with diverse visual concepts across domains. Among them, the CLIP series stands out as one of the most widely adopted paradigms ~\cite{cherti2023reproducible,sun2023eva}, providing a solid foundation for OVS due to its cross-modal capability.

Despite these advantages, CLIP is trained with a contrastive image-text objective that focuses on global alignment. As a result, it struggles to capture the fine-grained, pixel-level spatial details required for dense prediction tasks.
Recent studies have attempted to mitigate this issue by adopting a weakly supervised paradigm, relying solely on image–text pairs for OVS training ~\cite{wysoczanska2024clip,xu2022groupvit,luo2023segclip,wu2024image,wang2025dual}. At the same time, Vision Foundation Models (VFMs) ~\cite{caron2021emerging, he2022masked, oquab2024dinov2,kirillov2023segment} have emerged as exceptional visual encoders due to their superior fine-grained representations. In particular, the DINO series ~\cite{caron2021emerging, oquab2024dinov2} shows robust performance in spatial localization and semantic understanding. It has therefore become a common backbone in many OVS frameworks, helping bridge the gap between global semantics and dense features ~\cite{barsellotti2025talking, jose2025dinov2}.

Even with these improvements, DINO-based methods still struggle to produce accurate, boundary-aware features in complex scenes. This limitation mainly arises from its self-supervised training paradigm, which emphasizes high-level semantic consistency while overlooking fine structural details. As shown in Fig.~\ref{fig:performance}, these methods perform suboptimally on more complex benchmarks such as Cityscapes ~\cite{cordts2016cityscapes} and COCO-Stuff ~\cite{caesar2018coco}, which feature diverse scenes with a larger number of objects and require higher boundary precision, especially when delineating object boundaries.

We find through investigation that DINO is not entirely lacking boundary awareness. Instead, this capability gradually weakens as features pass through deeper transformer layers, where the model increasingly prioritizes abstract semantics. To compensate for this weakness, the Segment Anything Model (SAM) ~\cite{kirillov2023segment} provides strong boundary perception and contour modeling thanks to its instance segmentation training objective.
Nonetheless, directly combining DINO with SAM is non-trivial. Their training objectives differ significantly, and they operate at different resolutions, leading to a mismatch in representation spaces. Naively injecting SAM signals into DINO features may disrupt the semantic structure learned by DINO.

To tackle these challenges, we propose OVS-DINO, a framework that leverages SAM to enhance the boundary awareness of DINO while preserving its localization and cross-modal alignment strengths. Specifically, we introduce a Structure-Aware Encoder (SAE) that aggregates multi-layer DINO features and projects them into the SAM latent space for structural alignment. We then design a Structure-Modulated Decoder (SMD) to map the enhanced features back into the DINO semantic space. To preserve semantic consistency, particularly to protect the cross-modal semantic space of DINO, we further introduce a Preservation Gate. Finally, a lightweight projection layer maps the refined features into the text-aligned space for text-guided mask prediction.
During training, SAM latent features are used as structural alignment targets, while its generated pseudo-masks provide explicit supervision. Together, these signals ensure accurate and robust Open-Vocabulary Segmentation.

Extensive experiments show that OVS-DINO achieves superior performance across multiple benchmarks and consistently outperforms existing methods in complex scenarios. More importantly, it restores the boundary sensitivity of DINO, particularly addressing the structural degradation in deeper layers, and enables more precise contour prediction, leading to an average improvement of 2.1\% and notable gains of 6.1\% on Cityscapes and 6.3\% on COCO-Stuff.

In summary, our main contributions are as follows:
\begin{itemize}
\item We propose a novel architecture consisting of a Structure-Aware Encoder (SAE) and a Structure-Modulated Decoder (SMD). By leveraging SAM, our method enhances the boundary awareness of DINO and improves fine-grained structural representation.
\item We introduce a Preservation Gate to mitigate the feature discrepancy between SAM and DINO. This mechanism helps maintain the semantic space of DINO and preserves its cross-modal capabilities.
\item We achieve new \textbf{state-of-the-art} results on eight OVS benchmarks. Our method consistently outperforms existing weakly supervised approaches, especially in complex and cluttered scenarios.
\end{itemize}

\section{Related work}

\subsection{Vision-Language Pre-training}

Large-scale image-text data has enabled models to learn shared cross-modal semantics, moving beyond modality-specific representations. Contrastive dual-encoder frameworks have become dominant in this paradigm, with CLIP ~\cite{radford2021learning} as a representative example. Trained on web-scale data, CLIP demonstrates strong transferability and has been widely applied to classification, detection ~\cite{zhong2022regionclip,zhao2022exploiting}, and segmentation ~\cite{liang2023open,zhang2023uncovering}. However, its reliance on global image-level alignment limits the modeling of fine-grained spatial details, which is critical for dense prediction tasks. To address this, prior works either refine supervision from image-text pairs to encourage implicit region-level alignment ~\cite{wu2023clipself,cha2023learning}, or introduce additional grounding annotations for explicit region-text correspondence. Different from these approaches, we decouple visual and textual representation learning. We use CLIP text encoder to provide semantic guidance, while adopting DINO ~\cite{oquab2024dinov2} as the visual backbone, enabling alignment between textual features and fine-grained visual representations.

\subsection{Vision Foundation Models}

Recent progress in vision foundation models has been largely driven by advances in modern backbone architectures ~\cite{dosovitskiy2020image,he2016deep,o2015introduction} and large-scale pretraining. In particular, self-supervised learning (SSL) methods ~\cite{caron2021emerging,he2022masked,zhou2021ibot,bao2021beit} have demonstrated strong capability in capturing transferable semantics while preserving fine-grained spatial structures. Among them, the DINO family ~\cite{caron2021emerging,oquab2024dinov2} is known for producing spatially coherent representations that reflect the semantic layout of images ~\cite{darcet2023vision,simeoni2021localizing}. Complementary to representation learning, segmentation-oriented models such as SAM ~\cite{kirillov2023segment} leverage large-scale mask annotations to learn precise boundary-aware features. These models are particularly effective in capturing contour details and object-level structures. Our framework leverages the SAM features focus on edge contours to activate the DINO model ability, thus making it better suited for dense prediction tasks.

\subsection{Open-Vocabulary Segmentation}

Open-Vocabulary Segmentation (OVS) training methods can be divided into two categories: fully-supervised and weakly-supervised. Fully-supervised approaches leverage manual annotations to learn pixel-level alignment while expanding vocabulary through image-text supervision. SAN ~\cite{xu2023side} jointly performs mask generation and classification by introducing a side-adaptive network on a frozen CLIP, while ESC-Net ~\cite{lee2025effective} uses CLIP-derived pseudo cues to guide SAM for improved spatial aggregation and cross-modal interaction. Although these methods produce high-quality segmentation masks, their reliance on costly manual annotations limits scalability and restricts applicability to domains with available annotations.

Weakly-supervised methods aim to learn segmentation from image-text pairs without manual annotations. Early approaches focus on grouping visual tokens into semantic regions, such as GroupViT ~\cite{xu2022groupvit} and SegCLIP ~\cite{luo2023segclip}, which aggregate patches into region-level representations guided by text supervision. More recent works emphasize explicit region-text alignment. For example, TCL ~\cite{cha2023learning} adopts text-based contrastive learning, while CoDe ~\cite{wu2024image} decomposes images and text into regions and word-level units to strengthen cross-modal correspondence. CLIP-DINOiser ~\cite{wysoczanska2024clip} optimizes the dense features of CLIP by incorporating localization prior knowledge from DINO. DSGS ~\cite{wang2025dual} leverages SAM to produce structural semantic guidance and integrates it with textual semantics, forming a dual-guidance mechanism that supervises the CLIP model for improved performance in dense prediction tasks. Talk2DINO ~\cite{barsellotti2025talking} simply uses a projection function to map the CLIP text embedding into the DINOv2 space, and leverages self-attention to select key regions to achieve fine-grained cross-modal alignment. Compared with the above approaches, we replace the image encoder in CLIP with DINO to improve local feature quality. Furthermore, we refine the last layers of DINO using supervision derived from SAM, where pseudo-masks are used to guide training and enhance sensitivity to object boundaries.

\section{Preliminary}

\begin{figure}[t]
  \centering
  \includegraphics[width=0.4\textwidth]{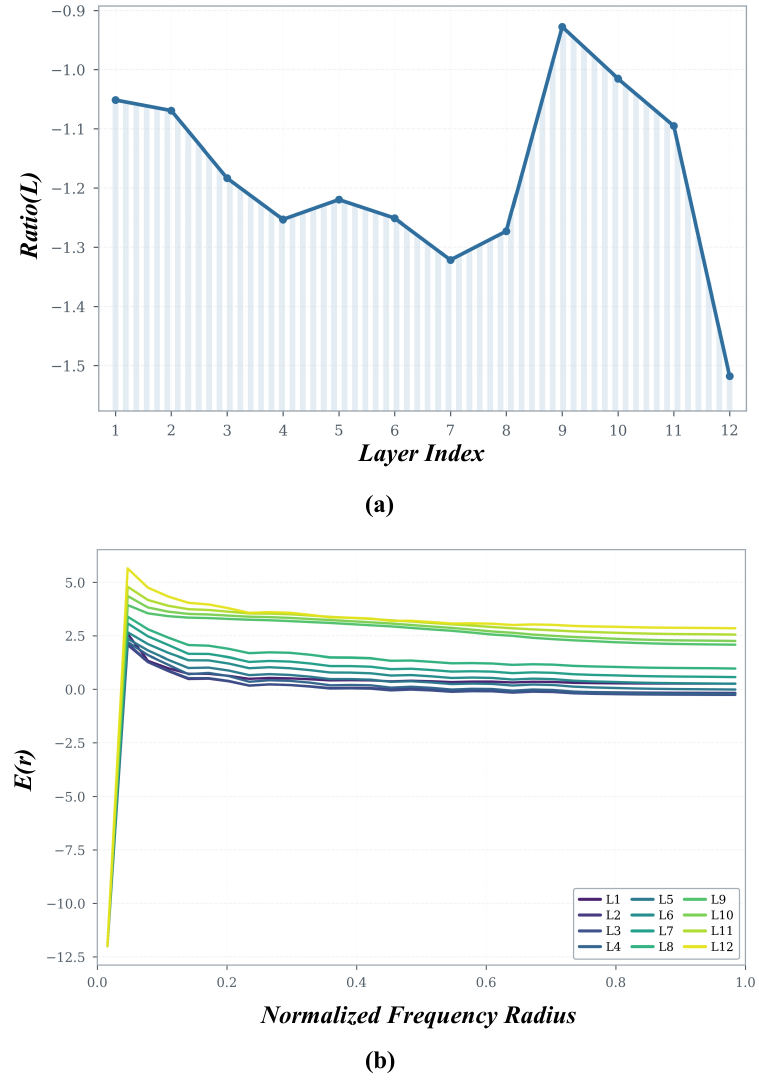}
  \caption{\textbf{Frequency-domain analysis of DINO layers.} (a) The high-to-low frequency energy ratio exhibits a consistent downward trend across layers, indicating a progressive transition from high-frequency local details to low-frequency global semantics during inference. (b) The azimuthally averaged power spectrum $E(r)$ shows that at higher frequencies (large $r$), deeper layers exhibit lower energy compared to shallow ones, indicating that shallow layers effectively preserve more fine-grained edge information.}
  \label{fig:math}
\end{figure}

\paragraph{Task Definition.} Open-Vocabulary Segmentation (OVS) based on Vision Transformers (ViTs) ~\cite{dosovitskiy2020image} follows a structured pipeline in which patch-level visual features are first extracted and treated as the fundamental representation units ~\cite{luo2023segclip}. Building upon these features, the model bridges the semantic space between dense visual representations and textual embeddings to achieve cross-modal alignment ~\cite{xu2022groupvit}. Specifically, the similarity between patch-level features and text category embeddings is computed to produce category-wise response maps, which are subsequently upsampled to the original image resolution, thereby enabling the assignment of semantic labels to each pixel. Formally, the overall process can be defined as follows: 

\begin{equation}
S_{h,w}^{(j)} = \frac{\mathbf{v}_{h,w} \cdot \mathbf{t}_j^{\top}}{\|\mathbf{v}_{h,w}\| \|\mathbf{t}_j\|},
\label{eq:patch_similarity}
\end{equation}

\begin{equation}
L_{h,w} = \arg\max_{j \in \{1, \dots, M\}} S_{h,w}^{(j)},
\label{eq:patch_assignment}
\end{equation}
where $S_{h,w}^{(j)}$ denotes the semantic similarity between the visual patch at $(h, w)$ and the $j$-th textual category, computed based on the visual feature $\mathbf{v}_{h,w} \in \mathbb{R}^{D_v}$ and the corresponding textual embedding $\mathbf{t}j \in \mathbb{R}^{D_t}$. Based on these similarity scores, the predicted semantic label at location $(h, w)$ is denoted as $L_{h,w}$.

\begin{figure}[t]
  \centering
  \includegraphics[width=0.45\textwidth]{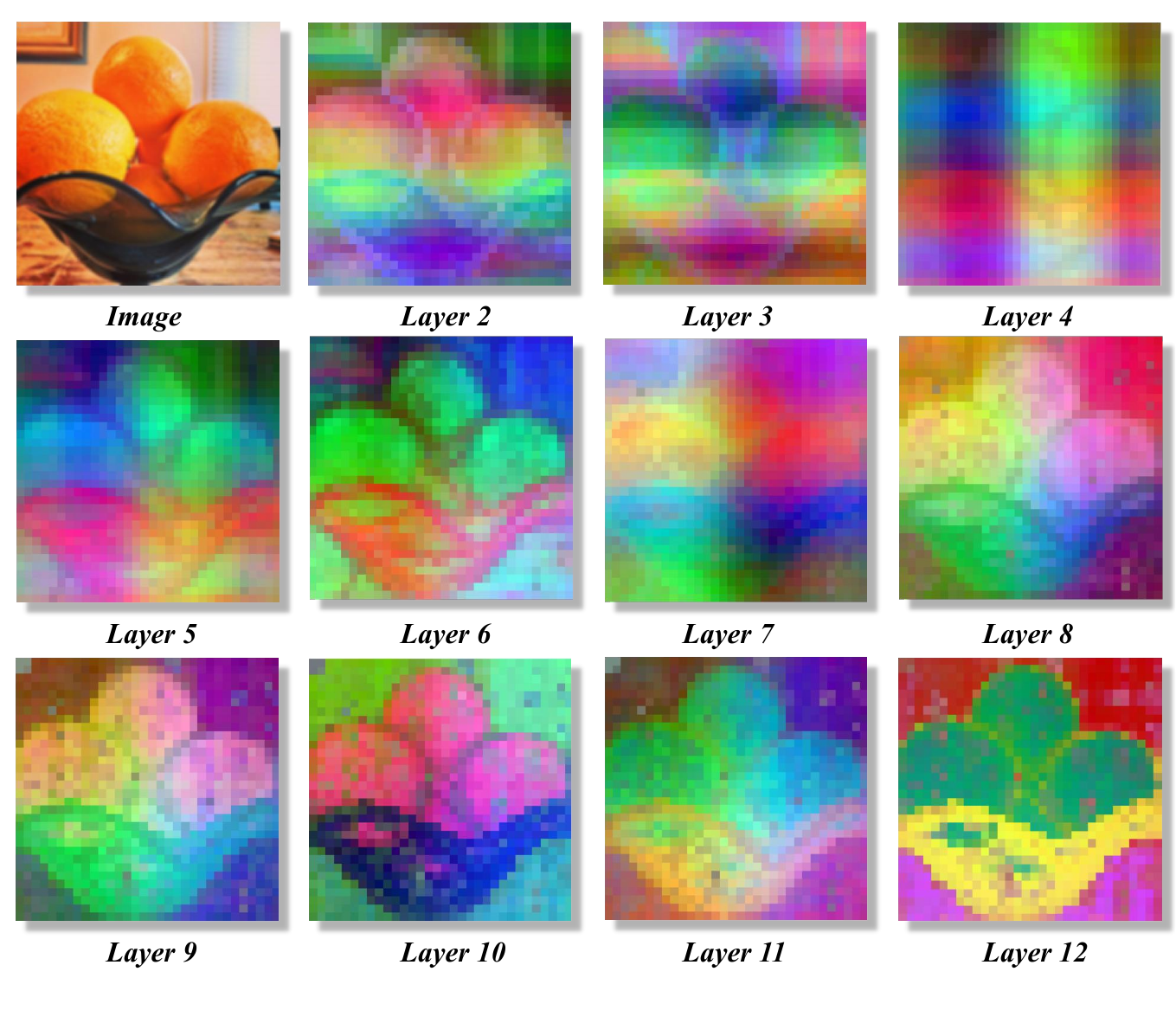}
  \caption{\textbf{Layer-wise PCA visualization.} DINO exhibits strong object contour awareness as early as its initial layers, indicating that structural information is captured during the early stages of feature extraction ~\cite{caron2021emerging}.}
  \label{fig:vis}  
\end{figure}

\paragraph{Analysis of DINO Space.} Under the aligned textual semantic space of CLIP, numerous previous works have demonstrated that DINO ~\cite{caron2021emerging,oquab2024dinov2}, as a self-supervised visual backbone, not only exhibits strong fine-grained representation capability but also maintains notable consistency and compatibility with the CLIP text embedding space ~\cite{oquab2024dinov2,barsellotti2025talking}. However, although these methods perform well in relatively simple segmentation scenarios (e.g., VOC21 ~\cite{everingham2010pascal} and VOC20 ~\cite{everingham2010pascal}), they struggle to produce precise boundary delineation in more complex multi-object settings, leading to suboptimal performance on challenging datasets such as Cityscapes ~\cite{cordts2016cityscapes} and COCO-Stuff ~\cite{caesar2018coco} (Fig.~\ref{fig:performance}).

As illustrated in the Fig.~\ref{fig:vis}, DINO is not inherently limited in its ability to capture fine-grained boundaries or handle complex scenes. Rather, due to its self-supervised training paradigm, which primarily emphasizes high-level semantic abstraction, such capabilities gradually diminish as the depth of the ViT layers increases~\cite{raghu2017svcca,kornblith2019similarity}.

In Fig.~\ref{fig:math}, following ~\cite{rahaman2019spectral}, we analyze the frequency characteristics of DINO features using the 2D Discrete Fourier Transform (DFT). Given a feature map $f(x, y)$ from layer $L$, its frequency representation is defined as:
\begin{equation}
F(u, v) = \sum_{x=0}^{M-1} \sum_{y=0}^{N-1} f(x, y) e^{-i 2\pi \left(\frac{ux}{M} + \frac{vy}{N}\right)},
\label{eq:dft}
\end{equation}
with the corresponding power spectrum:
\begin{equation}
P(u, v) = |F(u, v)|^2.
\label{eq:power}
\end{equation}
Here respectively, low high frequency components capture semantic structures and fine details.

To enable layer-wise comparison, we compute the radial energy profile via azimuthal averaging:
\begin{equation}
E(r) = \text{Avg}_{\sqrt{u^2 + v^2} = r} P(u, v),
\label{eq:radial}
\end{equation}
where $r$ denotes the normalized frequency radius. As shown in Fig.~\ref{fig:math}, $E(r)$ exhibits a clear decay pattern, with high-frequency energy progressively diminishing as layer depth increases.

To quantify this effect, we measure the log-ratio between high- and low-frequency energy:
\begin{equation}
\text{Ratio}(L) = \log_{10} \left( \frac{\int_{R_c}^{R_{\max}} E_L(r)\,dr}{\int_{0}^{R_c} E_L(r)\,dr} \right),
\label{eq:ratio}
\end{equation}
which decreases monotonically across layers, suggesting that DINO behaves as an implicit low-pass filter.

This spectral bias indicates that, although the space of DINO inherently preserves a certain degree of edge awareness, this capability progressively degrades with increasing depth, ultimately giving way to dominant semantic abstraction. Therefore, the key challenge lies in preserving the strong semantic representation of deep layers while reactivating their intrinsic fine-grained boundary sensitivity.

\begin{figure}[t]
  \centering
  \includegraphics[width=0.45\textwidth]{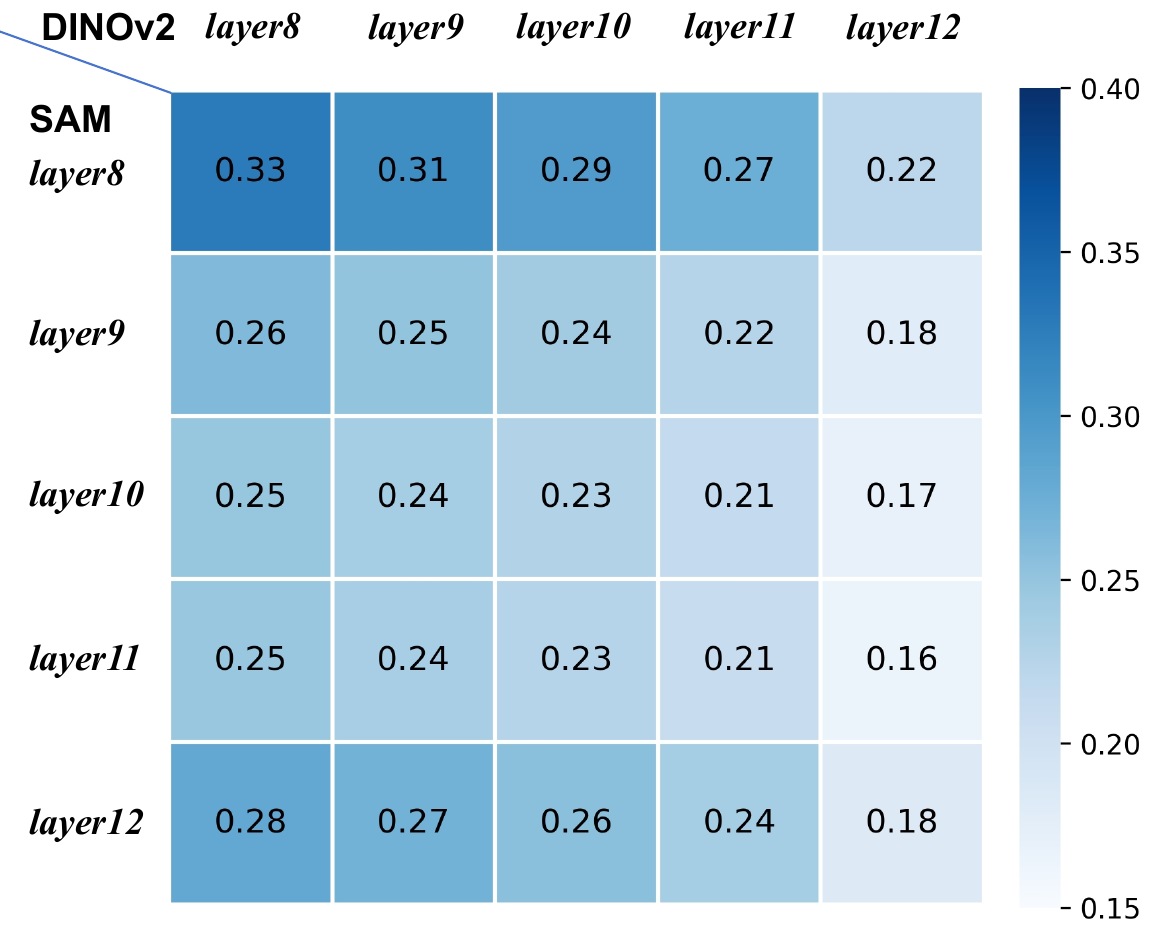}
  \caption{CKA heatmap results of DINOv2 and SAM, obtained from experiments conducted on the COCO dataset ~\cite{caesar2018coco}.}
  \label{fig:cka}  
\end{figure}

\paragraph{Analysis of SAM Space.}As a powerful instance segmentation model, SAM relies on a high-resolution image encoder coupled with a dedicated mask decoder for prediction ~\cite{kirillov2023segment}. Extensive prior works ~\cite{ke2023segment,tang2023can,ma2024segment} have shown that SAM possesses strong edge awareness and fine-grained representation capability, which are precisely the properties lacking in DINO. However, as a fully supervised model tailored for instance segmentation, the space of SAM is highly specialized to serve its decoder, resulting in a significant gap from the space of DINO (Fig.~\ref{fig:cka}). Following the method in ~\cite{kornblith2019similarity}, we conduct a CKA similarity analysis on the last five layers of DINOv2 and SAM. The diagonal values in the CKA matrix reflect the similarity between the two feature spaces, where values closer to 1 indicate higher similarity. This discrepancy makes SAM difficult to directly adapt to Open-Vocabulary Segmentation scenarios. This poses the challenge of preserving the intrinsic semantic consistency of DINOv2 while effectively transferring the fine-grained capabilities of SAM.

\paragraph{Summary.} In summary, while DINO establishes a robust semantic foundation, boundary awareness undergoes progressive attenuation as features transition to deeper layers. Furthermore, the inherent resolution mismatch ($224$ vs $1024$) and the substantial representation gap between SAM and DINO prevent straightforward integration, as naive fine-tuning often leads to the degradation of the intrinsic cross-modal consistency of the visual features. These observations motivate the proposed framework, which leverages structural priors from SAM to reactivate the latent edge-sensitivity of DINO while preserving the integrity of the semantic space.

\begin{figure*}[t]
  \centering
  \includegraphics[width=\textwidth]{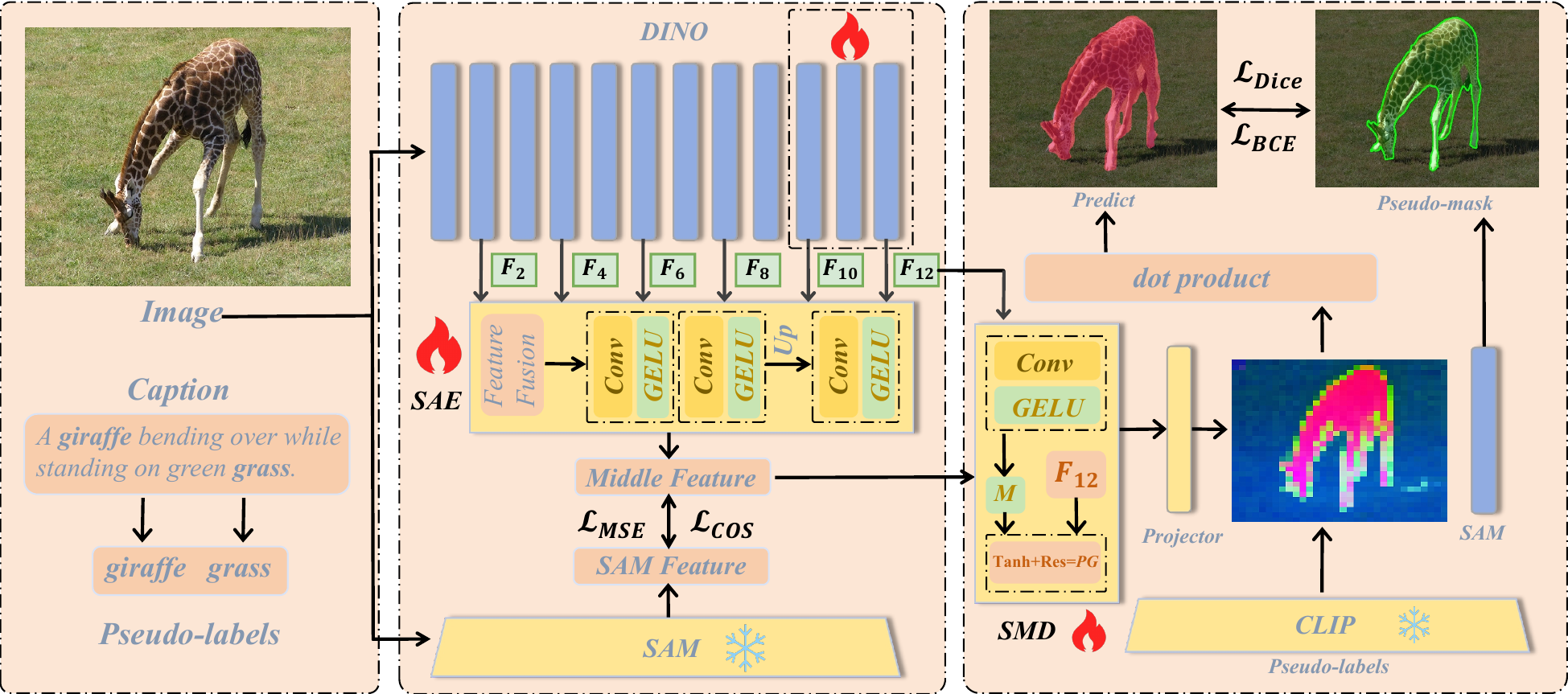}
  \caption{Overall architecture of OVS-DINO. We first encode the multi-level features from DINO using the Structure-Aware Encoder (SAE) and align them with SAM features. The aligned features are then passed through the Structure-Modulated Decoder (SMD) and projection layer to map them into the CLIP text space for subsequent segmentation. The alignment loss $\mathcal{L}_{\text{Align}} $ updates only the SAE, SMD, and projector, while the segmentation loss $\mathcal{L}_{\text{Seg}} $ further updates the last three layers of DINO in addition to these modules. For the training data, we use NLTK ~\cite{bird2009natural} to extract nouns from natural language descriptions, followed by refinement and filtering. Grounding DINO ~\cite{liu2024grounding} and SAM are then applied to generate the corresponding pseudo-masks. PG is an abbreviation for Preservation Gate.}
  \label{fig:pipeline}
\end{figure*}
\section{Proposed Method}


To resolve the tension between semantic abstraction and structural precision, we introduce OVS-DINO, a framework designed to reactivate the structural awareness embedded in shallow layers of DINO while preserving the deep semantic space. As illustrated in Fig.~\ref{fig:pipeline}, the architecture facilitates this through a Structure-Aware Encoder (SAE) and a Structure-Modulated Decoder (SMD). Specifically, the encoder integrates multi-level features of DINO into an intermediate representation, which is aligned with the feature space of SAM to capture fine-grained structural cues. Subsequently, the decoder maps this enhanced representation back to the original manifold of DINO. This process facilitates the incorporation of detailed texture and boundary information derived from SAM, thereby enhancing edge sensitivity without compromising the consistency of the cross-modal semantic space.

\subsection{Activate the edge awareness of DINO}
\paragraph{Structure-Aware Encoder (SAE)}
We first design a gated feature fusion module composed of a convolution layer and Group Normalization. This module normalizes the statistical properties across different layers and performs linear reorganization along the channel dimension, followed by weighted aggregation:
\begin{equation}
\hat{F} = F_{12} + \sum_{i \in \{2,4,6,8,10\}} \alpha_i \cdot G(F_i),
\end{equation}
where $G(\cdot)$ denotes the transformation function implemented by the fusion module.

To preserve the spatial structure and geometric information of DINO during encoding, we further employ two successive $3 \times 3$ depthwise convolution layers to refine and reorganize the fused features. After upsampling, a standard convolution layer is applied to produce the final representation, which is aligned with the feature space of SAM:
\begin{equation}
F_{\text{Middle}} = (\mathcal{L}_{d=2} \circ \mathcal{L}_{d=1})(\hat{F}),
\end{equation}
where each layer is defined as
\begin{equation}
\mathcal{L}_d = \text{GELU} ( \text{GroupNorm} \circ \text{Conv}_d).
\end{equation}

\paragraph{Structural Alignment.}
To effectively activate the edge awareness capability within $F_{\text{Middle}}$, we design an alignment objective using the features of SAM as a representational reference. This alignment ensures that the encoder internalizes the fine-grained structural priors of SAM. Specifically, we employ a combination of semantic and structural constraints. First, a cosine similarity loss is employed to enforce semantic alignment:
\begin{equation}
\mathcal{L}_{\text{cosine}} = 1 - \frac{F_{\text{Middle}} \cdot F_{\text{SAM}}}{\|F_{\text{Middle}}\|_2 \cdot \|F_{\text{SAM}}\|_2},
\end{equation}
which encourages consistency in the semantic space. Furthermore, to bridge the resolution gap and enable the low-resolution features of DINO to capture the fine-grained structural details from the high-resolution outputs of SAM, a mean squared error (MSE) loss is introduced:
\begin{equation}
\mathcal{L}_{\text{MSE}} = \frac{1}{d} \|F_{\text{Middle}} - F_{\text{SAM}}\|_2^2,
\end{equation}
where $d$ denotes the feature dimension. 
The overall alignment loss is defined as:
\begin{equation}
\mathcal{L}_{\text{Align}} = \mathcal{L}_{\text{cosine}} + \mathcal{L}_{\text{MSE}}.
\end{equation}
Through the optimization of $\mathcal{L}{\text{Align}}$, $F_{\text{Middle}}$ absorbs the fine-grained texture and boundary information embedded in the SAM space, facilitating the activation of latent structural knowledge.

\subsection{Preserve the semantic consistency of DINO }
\paragraph{Structure-Modulated Decoder (SMD)}
Having successfully activated the structural awareness via the encoder, it is imperative to project these enhanced features back onto the original manifold of DINO. This step is crucial to maintain the inherent cross-modal alignment and semantic robustness of the visual backbone while integrating the newly acquired structural cues.
To achieve this mapping, we propose a structure-modulated decoder. The process begins by compressing the encoder representation $F_{\text{Middle}}$ into a modulation map $M$ via a depthwise downsampling block:
\begin{equation}
M = \mathrm{GELU}\big( \mathrm{GroupNorm} \circ \mathrm{Conv}_{s=2}(F_{\text{Middle}}) \big).
\end{equation}
To ensure spatial consistency and preserve the semantic capability of the visual features, we introduce a novel \textbf{Preservation Gate}. This component consists of a residual block and a nonlinear activation. During decoding, the projected feature $F_{12}$ is first bilinearly interpolated to match the resolution of $M$, and then passed through the Preservation Gate to obtain the refined representation $F_{\text{out}}$:
\begin{equation}
F_{\text{out}} = F_{12} + \gamma \cdot \big( F_{12} \odot \tanh(M) \big),
\end{equation}
where $\odot$ denotes the Hadamard product and $\gamma$ is a learnable scaling parameter. Through this gating mechanism, the decoder effectively constrains the output to remain within the semantic space of DINO, preserving its expressive capacity while maximally retaining the edge-sensitive details.

\subsection{Training and inference}

\paragraph{Training.}
During training, we introduce a segmentation loss to explicitly enhance the segmentation capability of DINO and improve the alignment between visual features and the textual semantic space. Specifically, the intermediate representation $F_{\text{Middle}}$ is fed into the decoder to obtain the final feature $F_{\text{out}}$, which is used for segmentation.

The feature $F_{\text{out}}$ is projected into the CLIP-aligned embedding space via a $1 \times 1$ convolution layer, enabling mask prediction conditioned on text features. The predicted masks are supervised by pseudo-labels generated from SAM, optimized using a combination of Dice loss and binary cross-entropy (BCE) loss:

\begin{equation}
\mathcal{L}_{\text{Seg}} = \mathcal{L}_{\text{Dice}} + \mathcal{L}_{\text{BCE}}.
\end{equation}

Finally, the overall training objective is defined as the combination of the alignment loss and the segmentation loss:

\begin{equation}
\mathcal{L} = \mathcal{L}_{\text{Align}} + \mathcal{L}_{\text{Seg}}.
\end{equation}

\paragraph{Inference.}
At inference time, the pipeline consists of DINO, the Structure-Aware Encoder (SAE), the Structure-Modulated Decoder (SMD), and the CLIP text encoder. Specifically, visual features are first extracted by DINO and refined through the encoder-decoder architecture, and the resulting features are then projected into the CLIP-aligned space for mask prediction conditioned on text prompts. This design enables efficient inference while preserving both semantic consistency and structural sensitivity.

\section{Experiments}
\subsection{Experimental Setup}
\paragraph{Datasets.}
We evaluate our model on eight widely-used benchmarks. PASCAL VOC20 ~\cite{everingham2010pascal}, PASCAL Context59 ~\cite{mottaghi2014role}, COCO-Stuff ~\cite{caesar2018coco}, Cityscapes ~\cite{cordts2016cityscapes} and ADE20K ~\cite{zhou2019semantic} contain 20, 59, 171, 19 and 150 semantic categories without background class. Whlie PASCAL VOC ~\cite{everingham2010pascal} and PASCAL Context ~\cite{mottaghi2014role} have the “background” category(with 21 and 60 semantic categories). COCO-Object ~\cite{caesar2018coco} is composed of 80 different foreground object classes. While the PASCAL VOC and Context benchmarks primarily evaluate baseline segmentation on common objects in relatively straightforward contexts, COCO-based datasets and Cityscapes shift the focus toward complex multi-object scene parsing. ADE20K further extends this evaluation by challenging the scalability and fine-grained recognition of model within a vast and diverse semantic taxonomy.
\paragraph{Implementation Details.}
We use DINOv2 ViT-B/14 ~\cite{oquab2024dinov2} as the base model and SAM ViT-B ~\cite{kirillov2023segment} as the alignment model at the image end, with the CLIP ViT-B/16 model at the text end. The size of input images is \(1024 \times 1024\) for SAM and \(448 \times 448\) for DINOv2. We employ Grounding DINO ~\cite{liu2024grounding} to provide input hints for SAM to generate pseudo-masks for matching nouns generated by NLTK ~\cite{bird2009natural} and detailed dataset construction is provided in the Appendix. We train the model with a batch size of 64, and a learning rate of \(1 \times 10^{-4}\) for encoder, decoder and projector, and \(1 \times 10^{-5} \)for fine-tuning DINOv2 (last three layers) for a total of 25 epochs with a warmup epoch on the COCO Captions 2017 dataset ~\cite{lin2014microsoft}. The cosine schedule ~\cite{nichol2021improved} is adopted to adjust the learning rate, and the weight decay of the AdamW optimizer ~\cite{loshchilov2017decoupled} is set to \(1 \times 10^{-4}\). 
\paragraph{Evaluation Protocol.}
We conduct our evaluations using the MMSegmentation toolbox ~\cite{contributors2020mmsegmentation}. Input images are rescaled such that their shorter side is 448 pixels, and inference is performed with a sliding-window strategy using a stride of 224. For all benchmarks, textual inputs are constructed by combining class names with the standard ImageNet prompt templates ~\cite{radford2021learning}. Semantic segmentation performance is assessed using the mean Intersection over Union (mIoU) metric. 
\begin{table*}[t]
\centering
\small
\caption{Comparison of different methods across multiple datasets. We report results on benchmarks with background classes (VOC21 ~\cite{everingham2010pascal}, Context60 ~\cite{mottaghi2014role}, and Object ~\cite{caesar2018coco}) as well as without background classes (VOC20 ~\cite{everingham2010pascal}, Context59 ~\cite{mottaghi2014role}, etc.). We annotate the backbone used by each method and compare their average performance to reflect overall capability. Notably, our method achieves state-of-the-art performance on all four datasets as well as in terms of overall average.}
\resizebox{\textwidth}{!}{
\begin{tabular}{lcccccccccc}
\toprule
Model & Backbone &VOC21 & Context60 & Object & VOC20 & Context59 & Stuff & City & ADE & Avg \\
\midrule
GroupViT (CVPR2022 ~\cite{xu2022groupvit}) & CLIP S/16 & 50.4 & 18.7 & 27.5 & 79.7 & 23.4 & 15.3 & 11.1 & 9.2 & 29.4 \\
ReCo (NeurIPS2022~\cite{shin2022reco}) & CLIP B/16& 25.1 & 19.9 & 15.7 & 57.7 & 22.3 & 14.8 & 21.1 & 11.2 & 23.5 \\
TCL (CVPR2023~\cite{cha2023learning}) & CLIP B/16& 51.2 & 24.3 & 30.4 & 77.5 & 30.3 & 19.6 & 23.1 & 14.9 & 33.9 \\
CoDe (CVPR2024~\cite{wu2024image}) & CLIP B/16& 46.0 & 23.1 & 25.4 & 64.9 & 28.2 & 17.6 & 24.4 & 12.9 & 30.3 \\
CLIP-DINOiser (ECCV2024~\cite{wysoczanska2024clip}) & CLIP B/16& 62.1 & 32.4 & 34.8 & 80.9 & 35.9 & 24.6 & 31.1 & 20.0 & 40.2 \\
PixelCLIP-ViT (NeurIPS2024~\cite{shin2024towards}) & CLIP B/16& -- & -- & -- & 85.9 & 37.9 & 23.6 & 27.2 & 18.7 & -- \\
SynSeg (Arxiv2025)~\cite{zhang2025synseg}) & CLIP B/16& 62.2 & -- & 34.9 & -- & 41.8 & 23.6 & 30.9 & -- & -- \\
\texttt{dino.txt} (CVPR2025)~\cite{jose2025dinov2}) & DINOv2 L/14& -- & -- & -- & 62.1 & 30.9 & 20.9 & 32.1 & 20.6 & -- \\
Talk2DINO (ICCV2025~\cite{barsellotti2025talking}) & DINOv2 B/14 & 61.5 & 35.1 & \textbf{41.0} & \textbf{87.1} & 39.8 & 28.1 & 36.6 & \textbf{21.1} & 43.8 \\
DSGS (CVPR2025~\cite{wang2025dual}) & CLIP B/16& \textbf{67.7} & 40.8 & 36.3 & 85.0 & 46.5 & 25.7 & 36.0 & 20.7 & 44.8 \\
\textbf{OVS-DINO (Ours)} & DINOv2 B/14 & 64.0 & \textbf{42.0} & 38.6 & 83.5 & \textbf{49.4} & \textbf{34.2} & \textbf{42.9} & 20.6 & \textbf{46.9} \\
\bottomrule
\end{tabular}
}
\label{tab:results}
\end{table*}
\subsection{Comparison with the State of the Art}
Table~\ref{tab:results} presents a comparison between OVS-DINO and existing weakly supervised methods across eight zero-shot benchmarks. For methods that do not report results on all datasets (e.g., PixelCLIP ~\cite{shin2024towards}), we restrict comparisons to the available benchmarks for fairness.
Overall, OVS-DINO achieves state-of-the-art average performance, improving by +2.1\%. In particular, it sets new state-of-the-art results on Context60 (+1.2\%), Context59 (+2.9\%), COCO-Stuff (+6.1\%), and Cityscapes (+6.3\%).
Compared with representative methods, Talk2DINO ~\cite{barsellotti2025talking} enhances semantic representation by aligning multi-head attention features with text embeddings via contrastive learning, achieving strong performance on simpler datasets such as VOC, but lacking explicit modeling of fine-grained boundaries. DSGS ~\cite{wang2025dual}, on the other hand, improves performance by generating text-guided pseudo labels on the image side, which is also effective in relatively simple segmentation scenarios. 
In contrast, our method focuses on enhancing boundary-aware representations by aligning DINO features with structural priors from SAM, while using pseudo labels only as auxiliary supervision. This design allows the model to better capture fine-grained spatial details without sacrificing semantic consistency.
Notably, on COCO-Stuff, Cityscapes, and the Context series, OVS-DINO significantly outperforms prior methods, demonstrating stronger robustness and more accurate boundary delineation in cluttered scenes. Meanwhile, our method also achieves competitive performance on VOC benchmarks, with results within 3.6\% and 3.7\% of the best reported performance on VOC20 and VOC21, respectively.
On the ADE20K dataset, our method maintains strong performance, with only a marginal gap of 0.5\% to the state of the art, further demonstrating its generalization capability.

\subsection{Ablation Studies}

\begin{table}[h]
\centering
\small
\caption{Ablation study of the effect of designed model. SA and PG are abbreviations for SAM alignment and Preservation Gate.}
\resizebox{\columnwidth}{!}{
\begin{tabular}{lccccccccc}
\toprule
Method & V21 & C60 & Object & V20 & C59 & Stuff & City & ADE & Avg \\
\midrule
w/o SA      & 60.1 & 40.0 & 38.3 & 75.1 & 47.4 & 32.9 & 39.4 & 19.7 & 44.1 \\
w/o Encoder & 55.2 & 36.5 & 32.5 & 72.2 & 42.6 & 27.6 & 37.7 & 18.3 & 40.3 \\
w/o PG      & 37.4 & 31.0 & 26.5 & 49.4 & 36.9 & 25.9 & 38.3 & 16.5 & 32.7 \\
Ours        & 64.0 & 42.0 & 38.6 & 83.5 & 49.4 & 34.2 & 42.9 & 20.6 & 46.9 \\
\bottomrule
\end{tabular}
}
\label{tab:ablation of designed}
\end{table}
\paragraph{Effect of SAM Guidance and DINO Representation.}
To further investigate the roles of SAM features and DINO representations in our method, we conduct a series of ablation studies in Table~\ref{tab:ablation of designed},articularly in the first and second rows.
We first remove the SAM branch and allow DINO features to be directly processed by the encoder–decoder for subsequent training. As shown in the first row, the absence of SAM feature guidance leads to consistent performance degradation across all datasets, with the most significant drop observed on VOC20. We attribute this decline to the lack of fine-grained supervision on object boundaries that SAM features provide. This result highlights that SAM guidance enhances the model’s ability to capture precise edges, which not only improves robustness in complex scenes but also boosts segmentation accuracy across diverse scenarios.
In addition, we remove both DINO and the encoder, using only SAM features fed into the decoder for downstream tasks. As shown in the second row, performance drops substantially across all benchmarks. Such performance degradation underscores that SAM features alone lack the semantic richness provided by DINO for effective cross-modal segmentation.
Taken together, these results confirm that DINO provides the essential semantic foundation for cross-modal understanding, while SAM contributes complementary structural cues. This validates our design choice of jointly leveraging both representations within a unified framework.

\paragraph{Effect of Preservation Gate.}
To further investigate the necessity of preserving the DINO feature space during activation, as well as the effectiveness of the proposed Preservation Gate, we refer to the ablation results in the third row of Table~\ref{tab:ablation of designed}. Specifically, we remove the Preservation Gate and directly feed the intermediate features into the decoder’s convolutional layers for mask prediction.
As evidenced by the data, eliminating the Preservation Gate leads to a severe performance drop across all benchmarks, with the most significant degradation observed among all ablation settings. This indicates that directly transforming intermediate features without preserving the original DINO semantic space substantially harms the model’s representation capability.
Without explicit preservation, the semantic consistency required for cross-modal understanding is compromised, leading to inferior segmentation performance.
In summary, these results highlight that preserving the DINO feature space is essential for maintaining semantic consistency during structural enhancement, validating the role of the Preservation Gate in enabling stable and effective feature integration.

\begin{table}[h]
\centering
\small
\caption{Ablation study of the layers selection and fine-tuning strategy.}
\resizebox{\columnwidth}{!}{
\begin{tabular}{lccccccccc}
\toprule
Method & V21 & C60 & Object & V20 & C59 & Stuff & City & ADE & Avg \\
\midrule
four layers & 61.2 & 41.0 & 38.9 & 80.2 & 48.7 & 34.4 & 41.4 & 19.8 & 45.7 \\
frozen DINO & 56.3 & 37.0 & 33.8 & 73.4 & 43.5 & 28.2 & 40.9 & 18.1 & 41.4 \\
Ours & 64.0 & 42.0 & 38.6 & 83.5 & 49.4 & 34.2 & 42.9 & 20.6 & 46.9 \\
\bottomrule

\end{tabular}
}
\label{tab:ablation of layers}
\end{table}
\begin{figure*}[t]
  \centering
  \includegraphics[width=\textwidth]{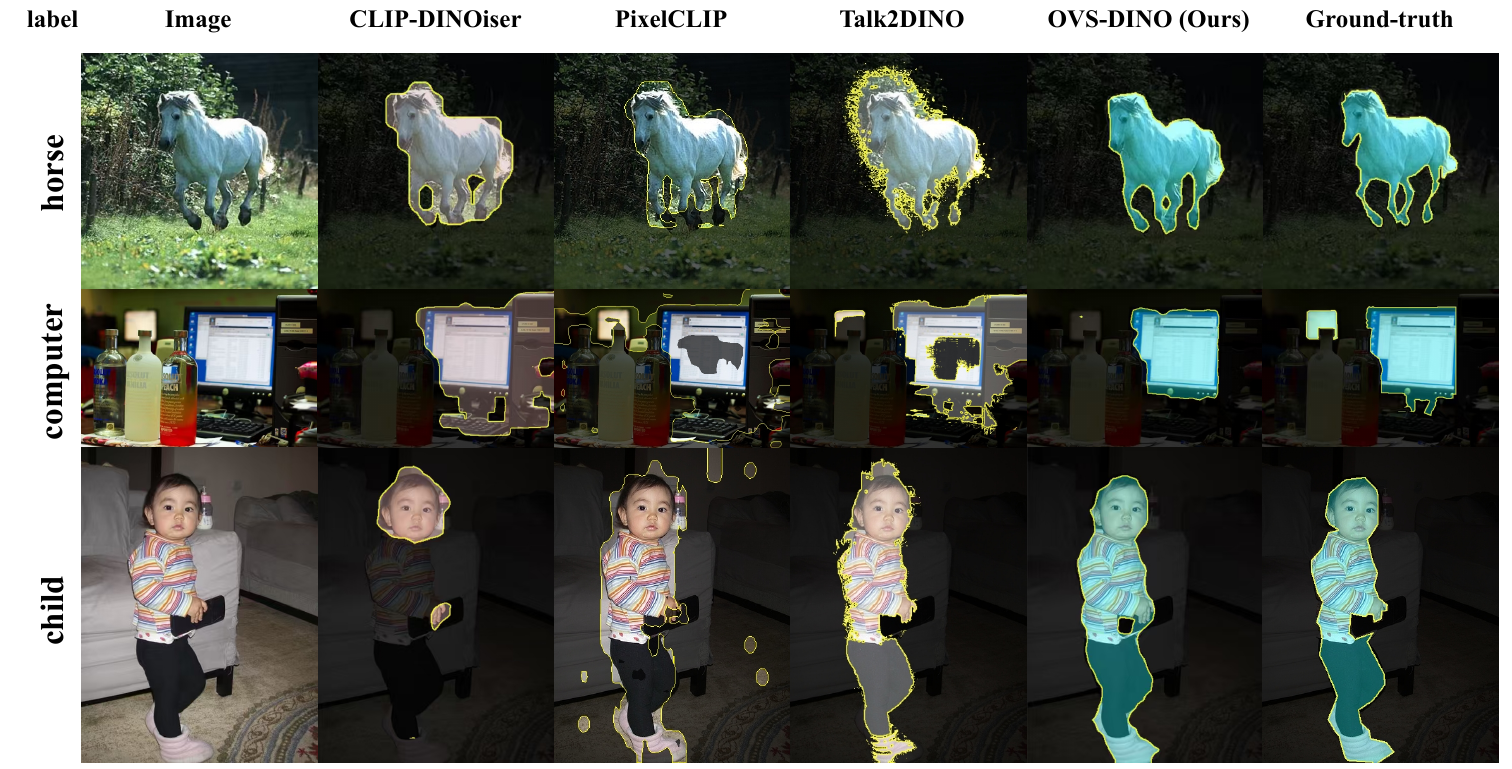}
  \caption{Visualization results of different models.}
  \label{fig:vis_results}
\end{figure*}
\paragraph{Impact of Layer Selection.}
We first investigate the effect of layer selection by modifying the input features to the encoder (Table~\ref{tab:ablation of layers}). Specifically, we replace the original dense sampling from layers \{2,4,6,8,10,12\} with a sparser configuration \{3,6,9,12\}. This change leads to a consistent performance drop across all benchmarks, 
which stems from the reduced ability to capture fine-grained structural information from shallow layers. Dense sampling of early features is critical for preserving high-frequency details, which aligns with our earlier analysis that boundary-aware cues are primarily encoded in shallow layers of DINO.

\paragraph{Analysis of Fine-tuning Strategy.} 
Furthermore, we explore whether the activation process can rely solely on the encoder–decoder by fully freezing DINO during training. Table~\ref{tab:ablation of layers} highlights the significant performance degradation caused by this setting. This observation indicates that structural guidance from SAM alone is insufficient to fully activate the representation capability of DINO. Instead, adapting the high-level semantic features of DINO through fine-tuning is essential for effective knowledge transfer and cross-modal alignment.
Overall, these findings indicate that both dense shallow-layer aggregation and partial fine-tuning are essential for effectively activating and adapting the DINO feature space under structural supervision.

\paragraph{Validation of Activation from SAM}
To further investigate the roles of SAM and DINO features in the segmentation pipeline, we modify the inference path without additional training. Specifically, instead of using the encoded DINO features for downstream prediction, we directly replace them with SAM features as input to the decoder and projector.
As shown in Table~\ref{tab:ablation of inference}, the performance remains nearly identical under both settings. This observation suggests that the decoder and projector have already learned how to effectively utilize structure-aware representations during training, regardless of whether the input originates from SAM or DINO.
This behavior can be explained by the implicit knowledge transfer introduced by the alignment objective. Through training, the structural priors from SAM are progressively incorporated into the DINO feature space, enabling the refined DINO representations to internalize boundary-aware information. As a result, the model no longer relies on explicit SAM features during inference.
This finding indicates that SAM primarily serves as a structural supervisor during training rather than a necessary component at inference time. Consequently, our method achieves comparable performance without incurring the significantly higher computational cost of SAM, making it more efficient for real-world deployment.
\begin{table}[h]
\centering
\small
\caption{Ablation study of the validation of activation from SAM.}
\resizebox{\columnwidth}{!}{
\begin{tabular}{lccccccccc}
\toprule
Feature & V21 & C60 & Object & V20 & C59 & Stuff & City & ADE & Avg \\
\midrule
SAM & 63.93 & 41.96 & 38.62 & 83.49 & 49.35 & 34.16 & 42.9 & 20.64 & 46.88 \\
DINO & 63.95 & 41.97 & 38.62 & 83.51 & 49.35 & 34.16 & 42.88 & 20.64 & 46.89 \\
\bottomrule
\end{tabular}
}
\label{tab:ablation of inference}
\end{table}
\subsection{Visualization Results}

In Fig.~\ref{fig:vis_results}, we present qualitative comparisons of mask predictions across different methods under various scenarios, including simple single-object scenes, multi-object settings, and human-centric segmentation tasks. 
Our method consistently produces more precise and coherent boundaries across all scenarios, with predictions that are visually closer to the ground-truth masks. Notably, in complex multi-object scenes (second row), our approach demonstrates strong localization capability, accurately separating multiple instances without confusion.
Furthermore, in challenging cases such as human segmentation, our method avoids over-segmentation, successfully capturing the complete human body rather than fragmenting it into partial regions. 
These results highlight the effectiveness of our approach in capturing fine-grained boundary details, while also demonstrating strong generalization and robustness in diverse scenarios.

\section{Conclusion}
We propose OVS-DINO, an Open-Vocabulary Segmentation framework that enhances boundary awareness in DINO via structural alignment with SAM. By mitigating detail loss from semantic abstraction, we introduce a Structure-Aware Encoder (SAE), a Structure-Modulated Decoder (SMD), and a Preservation Gate to fuse structural priors with semantic consistency. Our joint training strategy achieves state-of-the-art performance across benchmarks by activating latent edge sensitivity. Furthermore, transferring structural knowledge from SAM to DINO enables efficient inference without any reliance on SAM, while maintaining strong generalization and precise boundary delineation.
\bibliographystyle{ACM-Reference-Format}
\bibliography{sample-base}

\clearpage
\appendix
\setcounter{figure}{0}
\setcounter{table}{0}

\section{Dataset Creation Process}

\begin{strip}
    \centering
    \includegraphics[width=\textwidth]{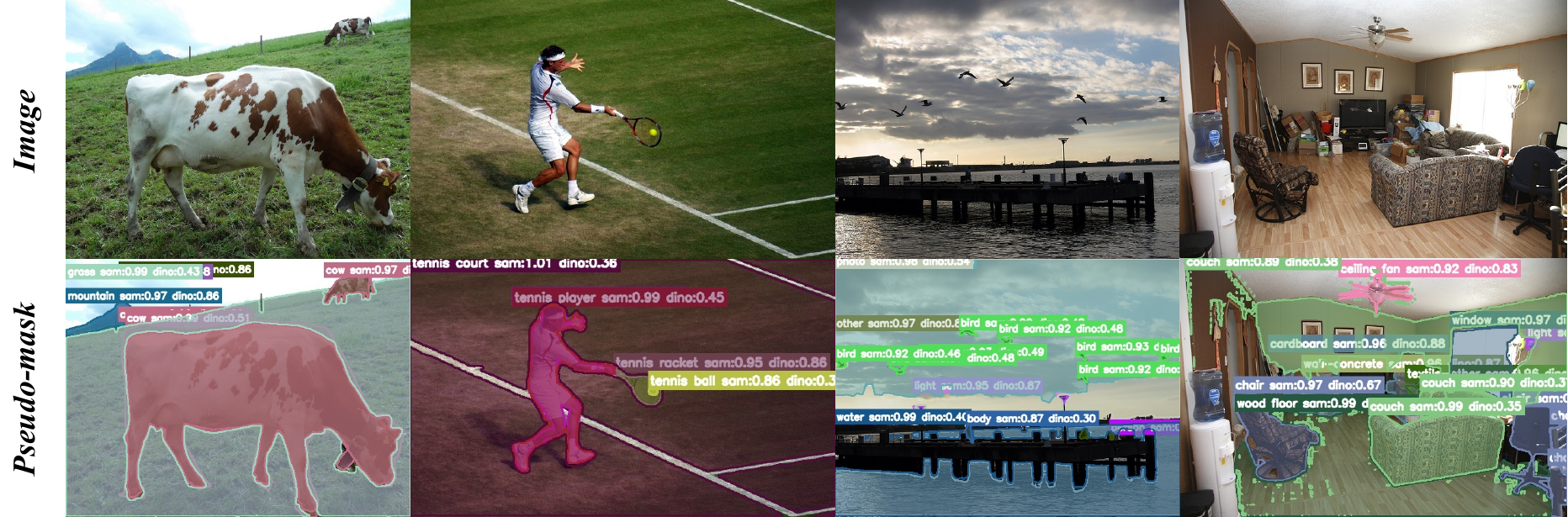}
    \captionof{figure}{Visualization of pseudo-masks.}
    \label{fig:pseudo-mask}
\end{strip}

Currently, many pseudo-mask generation methods rely on the panoptic segmentation mode of the SAM ~\cite{kirillov2023segment} model for automatic mask generation. However, this approach ~\cite{wang2025dual} utilizes uniformly sampled points as prompts, which leads to relatively slow inference speed and severe over-segmentation. Such over-segmentation is particularly detrimental to semantic segmentation tasks. In addition to point-based prompts, SAM also supports bounding box prompts. When a bounding box that accurately encloses an object is provided, the model can generate a more complete and coherent object mask, thereby effectively alleviating the over-segmentation problem. Motivated by this advantage, we adopt a pipeline that combines Grounding DINO ~\cite{liu2024grounding} and SAM to generate pseudo-masks. Grounding DINO is an open-vocabulary object detection model based on vision-language alignment. It is capable of localizing arbitrary semantic objects in an image using natural language descriptions, without being constrained by predefined category labels. The qualitative results of the generated pseudo-masks are illustrated in Fig. ~\ref{fig:pseudo-mask}. The pseudo-mask generation process consists of the following steps: 
\begin{itemize}
\item We employ NLTK ~\cite{bird2009natural} to extract noun categories from the dataset captions. NLTK is a widely used natural language processing toolkit that provides fundamental text processing functionalities, including tokenization, tagging, and syntactic analysis.
\item The extracted noun categories are filtered to remove infrequent, incomplete, or redundant entries.
\item The refined category set, together with the corresponding images, is fed into Grounding DINO to obtain object detection bounding boxes.
\item The detected bounding boxes for each category are then used as prompts for SAM to generate the final pseudo-masks, thereby aligning each textual category with its corresponding mask.
\end{itemize}

\section{Additional Ablation Study}

In Table. ~\ref{tab:ablation} we demonstrate the performance of our approach as the DINO ~\cite{caron2021emerging,oquab2024dinov2} visual backbone changes and all backbones use the ViT-B size. As shown in table, replacing DINO with DINOv2 consistently improves performance across all benchmarks. The consistent gains brought by DINOv2 highlight the importance of high-quality visual pretraining. Compared to DINO, DINOv2 produces more robust and transferable features, which better align with the requirements of Open-Vocabulary Segmentation tasks. We further analyze the impact of registers in DINOv2 by comparing its variants with and without registers as shown in second and third column. Registers are designed to mitigate artifacts in ViT feature maps, where certain tokens exhibit disproportionately large norms and lose spatial specificity. While such a mechanism can help stabilize representations and improve feature organization, our results show that its effectiveness is dataset-dependent. In particular, although registers can enhance the quality of attention maps in some cases, they may also introduce additional constraints that limit flexibility in others. This leads to a trade-off between structured representations and adaptability, resulting in complementary performance across different benchmarks. Considering its slightly better overall performance and greater flexibility, we adopt DINOv2 without registers as the default backbone in experiments.

\begin{figure*}[t]
    \centering
    \includegraphics[width=\textwidth]{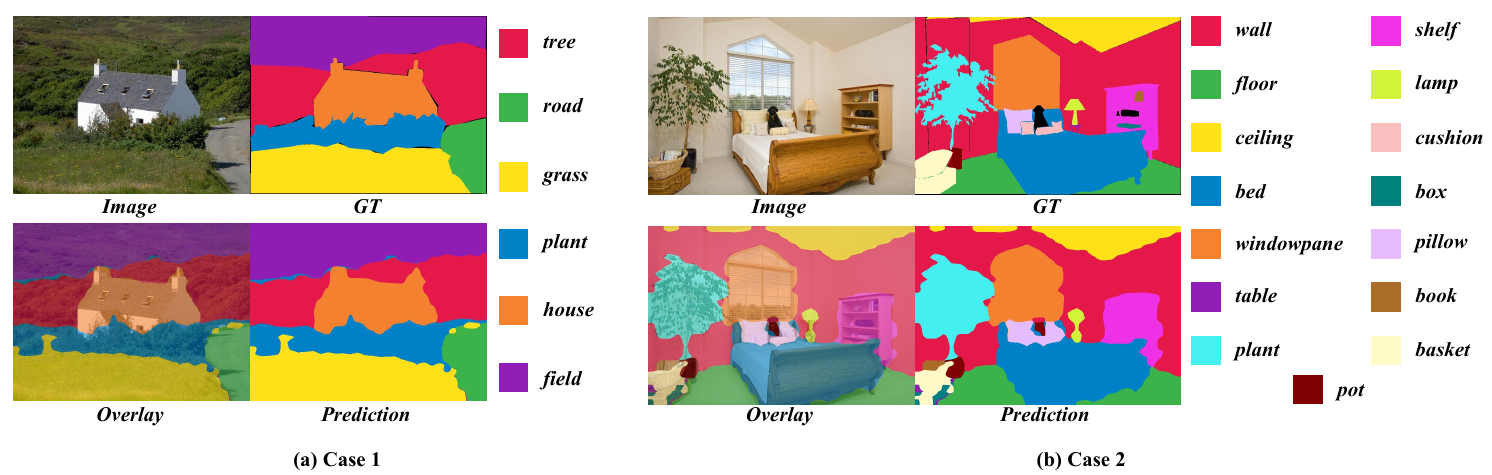}
    \caption{Visualization of success cases.}
    \label{fig:case12}
\end{figure*}

\begin{table}[h]
\centering
\small
\caption{Ablation study of using different visual backbones.}
\resizebox{\columnwidth}{!}{
\begin{tabular}{lccc}
\toprule
Dataset & DINO & DINOv2 (reg) & DINOv2 (Ours) \\
\midrule
VOC21    & 35.1 & 62.9 & 64.0 \\
Context60     & 26.6 & 41.9 & 42.0 \\
Object & 18.4 & 39.4 & 38.6 \\
VOC20    & 46.5 & 82.5 & 83.5 \\
Context59    & 30.3 & 49.7 & 49.4 \\
Stuff  & 18.2 & 34.8 & 34.2 \\
Cityscapes   & 27.6 & 41.8 & 42.9 \\
ADE20K    & 10.6 & 20.4 & 20.6 \\
Avg    & 28.7 & 46.7 & 46.9 \\
\bottomrule
\end{tabular}
}
\label{tab:ablation}
\end{table}

\section{Case Study}
\subsection{Success Case Analysis}


To qualitatively evaluate the effectiveness of the proposed model, Fig.~\ref{fig:case12} presents two representative successful cases from outdoor and indoor scenes. These examples demonstrate that the model is capable of producing accurate and semantically consistent predictions under the open-vocabulary setting, particularly for dominant regions and categories with distinctive visual characteristics.

In Case 1, the outdoor scene contains several semantically distinct categories, including house, tree, road, grass, and field. The model achieves high consistency with the ground truth in both region coverage and object boundaries. In particular, the structural outline of the house is well preserved, and clear separation is maintained between adjacent regions such as grass and road. This suggests that categories with distinctive visual patterns can be effectively aligned with their corresponding semantic representations in the shared embedding space. In addition, the relatively simple scene composition and limited category overlap further facilitate reliable visual-semantic matching. In Case 2, the indoor scene involves a larger number of categories, such as wall, floor, ceiling, bed, table, lamp, and windowpane, leading to increased scene complexity. Despite this, the model still produces accurate segmentation results for dominant regions, including walls, floors, and beds, while preserving the overall semantic structure of the scene. This indicates that the model is able to leverage semantically aligned visual representations conditioned on category descriptions, enabling consistent predictions across co-occurring objects within a unified embedding space.

Overall, these successful cases demonstrate that the proposed model performs reliably when sufficient visual-semantic cues are available, particularly for large regions and visually distinctive categories. The predictions exhibit strong global consistency and coherent scene structure, highlighting the model’s capability in capturing high-level semantic information under the open-vocabulary setting.

\subsection{Failure Case Analysis}

\begin{figure*}[t]
    \centering
    \includegraphics[width=\textwidth]{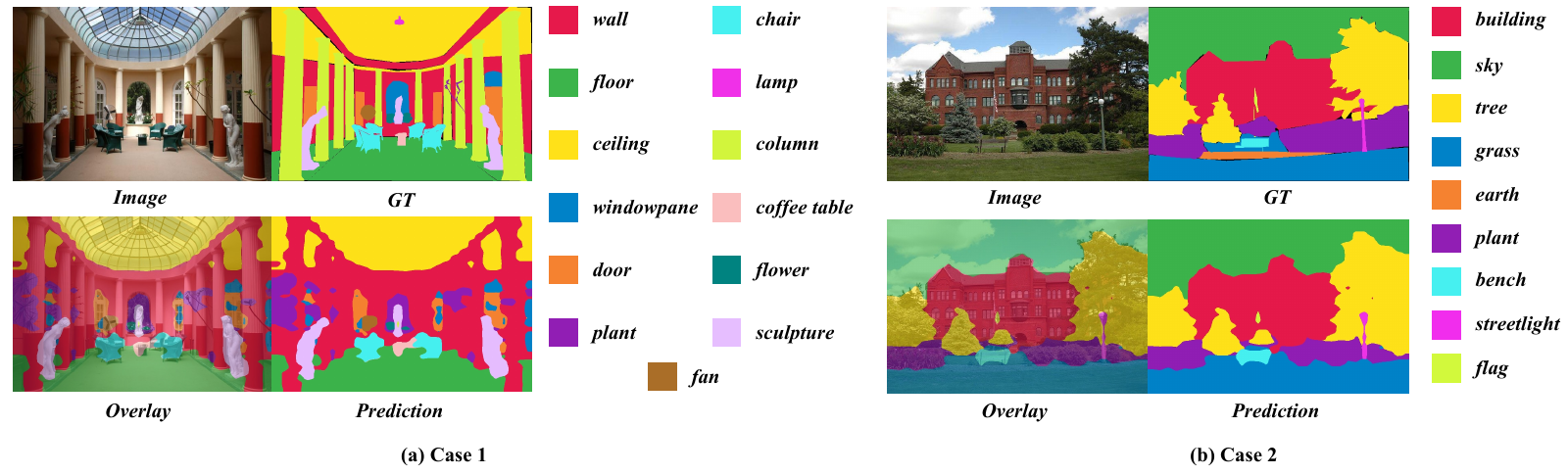}
    \caption{Visualization of failure cases.}
    \label{fig:case34}
\end{figure*}

To further investigate the limitations of the proposed model, Fig.~\ref{fig:case34} presents two representative failure cases from indoor and outdoor scenes. Compared with the successful examples, these cases show that our model still faces challenges in cluttered scenes, ambiguous category distinctions, and the segmentation of small or visually inconspicuous objects under the open-vocabulary setting.

\begin{figure*}[!b]
    \centering 
    \includegraphics[width=0.9\textwidth]{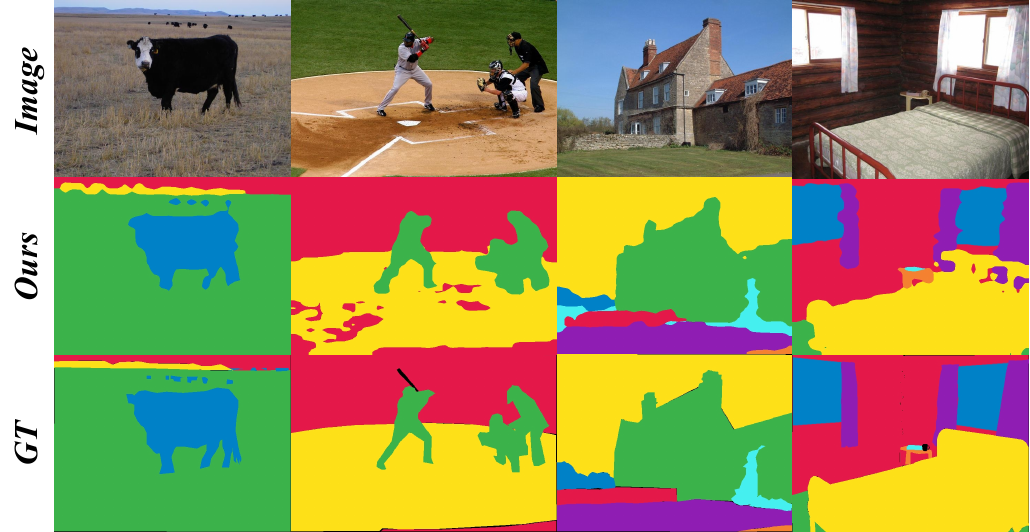} 
    \caption{Additional qualitative results on the ADE20K ~\cite{zhou2019semantic} dataset.} 
    \label{fig:ade} 
\end{figure*}

In Case 1, the indoor scene contains multiple semantic categories with substantial visual similarity and complex composition, including wall, floor, ceiling, windowpane, door, plant, chair, lamp, column, and sculpture. Although our method correctly identifies several dominant regions, such as the floor and part of the ceiling, noticeable discrepancies remain between the prediction and the ground truth. In particular, large portions of the scene are incorrectly assigned to the wall category, leading to confusion with structurally or semantically related regions such as columns, doors, and windowpanes. This suggests that, in complex indoor environments, categories with similar appearance or related semantics may not be sufficiently separated in the shared visual-semantic embedding space. In addition, thin or small-scale objects, such as lamps, flowers, and sculptures, are often missed or absorbed into surrounding regions, indicating limited robustness in aligning weak local visual evidence with fine-grained semantic concepts. Boundary ambiguity is also evident in densely structured areas, where the prediction tends to lose precise transitions between adjacent categories.

In Case 2, the model captures the overall semantic layout of the outdoor scene, particularly for large regions such as building, sky, and grass. However, several errors remain. The predicted boundaries are relatively coarse, especially around the building contour and tree canopy, resulting in the loss of structural details. This indicates that, although our approach can preserve global semantic consistency, it is less effective in distinguishing fine-grained structures when semantic evidence is locally ambiguous. Moreover, small categories such as streetlight and flag are poorly segmented, and some foreground regions are only partially recognized. These results suggest that large objects provide stronger and more stable visual-semantic cues, whereas small or less salient categories are more difficult to localize and classify accurately in an open-vocabulary setting.

Overall, these failure cases indicate that the proposed model performs reliably on dominant semantic regions but remains less robust for small objects, thin structures, and boundary-sensitive areas. In addition, visually or semantically similar categories are more likely to be confused, especially in cluttered indoor scenes. The predictions also exhibit a tendency toward over-smoothing in challenging regions, reflecting a trade-off between global semantic coherence and spatial precision. These observations highlight that, despite promising open-vocabulary segmentation capability, model still has limited fine-grained discrimination ability in complex scene parsing. Future improvements may focus on enhancing category separation in the shared embedding space, strengthening multi-scale representation for small objects and thin structures, and improving boundary-aware prediction under ambiguous visual conditions.

\section{Additional Qualitative Results}

\begin{figure*}[t]
    \centering
    \includegraphics[width=\textwidth]{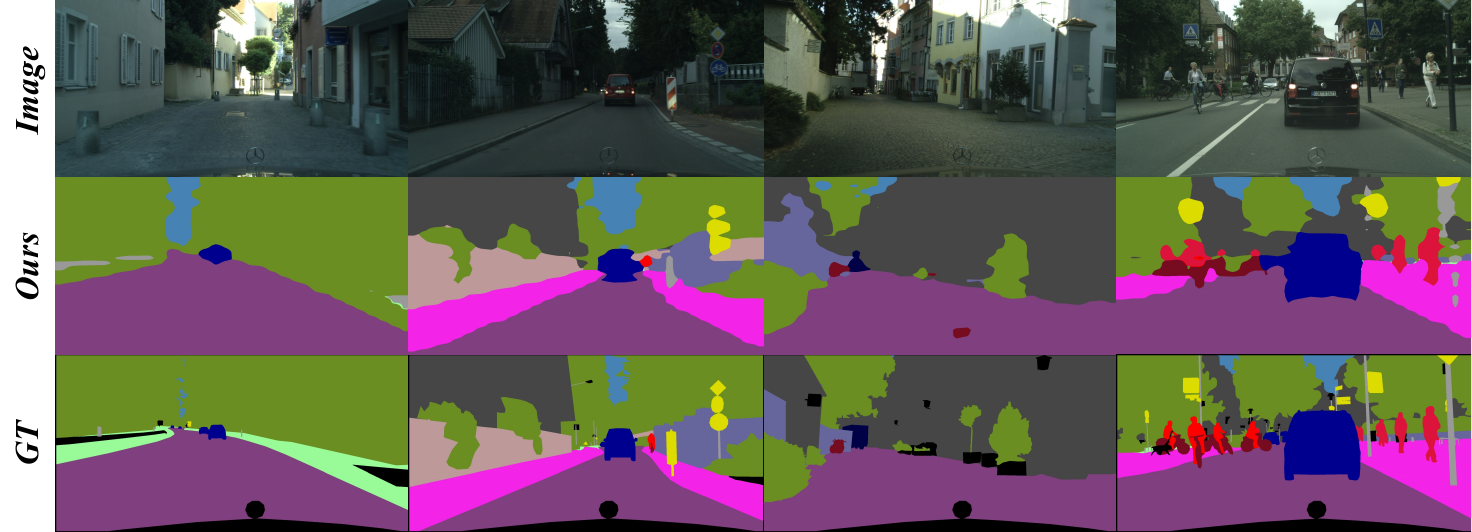}
    \caption{Additional qualitative results on the Cityscapes ~\cite{cordts2016cityscapes} dataset.}
    \label{fig:city}
\end{figure*}

\begin{figure*}[t]
    \centering
    \includegraphics[width=\textwidth]{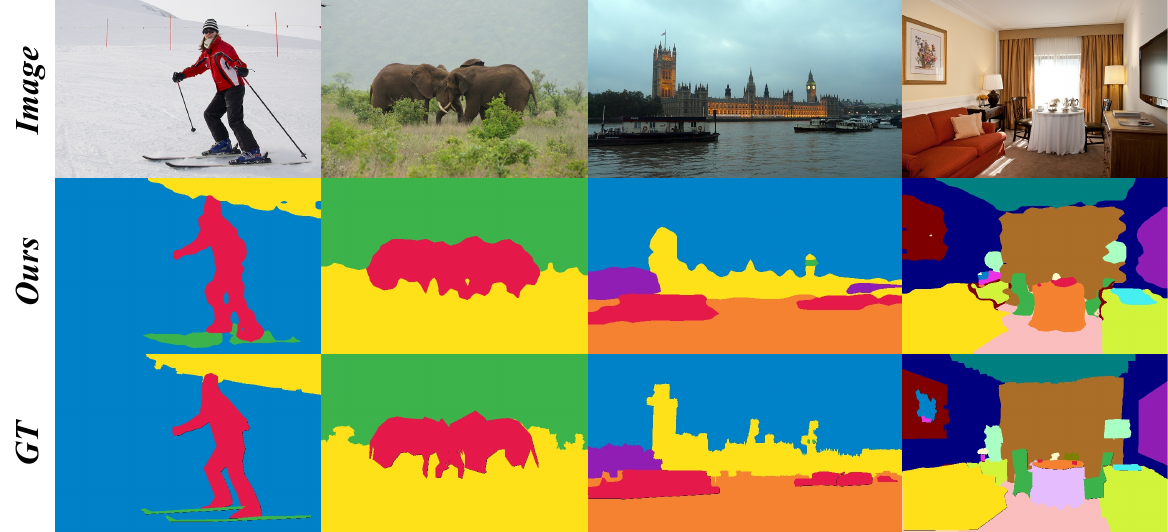}
    \caption{Additional qualitative results on the COCO-Stuff ~\cite{caesar2018coco} dataset.}
    \label{fig:stuff}
\end{figure*}

We present additional qualitative results on three datasets: ADE20K ~\cite{zhou2019semantic}, Cityscapes ~\cite{cordts2016cityscapes} and COCO-Stuff ~\cite{caesar2018coco}, as shown in Figs. ~\ref{fig:ade} to ~\ref{fig:stuff}. In Fig. ~\ref{fig:ade}, our method effectively segments regions across diverse categories, demonstrating robustness in both simple and complex scenarios. Furthermore, Figs. ~\ref{fig:city} and ~\ref{fig:stuff} highlight the performance of our approach on challenging multi-object scenes from Cityscapes and COCO-Stuff. Notably, OVS-DINO achieves relatively accurate segmentation of multiple objects even in highly cluttered environments.

\end{document}